%% file: tmlr.tex
\documentclass[10pt]{article} 
\usepackage[preprint]{tmlr}

\input{math_commands.tex}

\usepackage{hyperref}
\usepackage{url}

\usepackage{graphicx}

\title{Generative AI Systems: A Systems-based Perspective on Generative AI}

\author{\name Jakub M. Tomczak \email j.m.tomczak@tue.nl \\
      \addr Department of Mathematics and Computer Science\\
      Eindhoven University of Technology
      }

\def\month{MM}  
\def\year{YYYY} 
\def\openreview{\url{https://openreview.net/forum?id=XXXX}} 

\begin{document}

\maketitle

\begin{abstract}
Large Language Models (LLMs) have revolutionized AI systems by enabling communication with machines using natural language. Recent developments in Generative AI (GenAI) like Vision-Language Models (GPT-4V) and Gemini have shown great promise in using LLMs as multimodal systems. This new research line results in building Generative AI systems, GenAISys for short, that are capable of multimodal processing and content creation, as well as decision-making. GenAISys use natural language as a communication means and modality encoders as I/O interfaces for processing various data sources. They are also equipped with databases and external specialized tools, communicating with the system through a module for information retrieval and storage. This paper aims to explore and state new research directions in Generative AI Systems, including how to design GenAISys (compositionality, reliability, verifiability), build and train them, and what can be learned from the system-based perspective. Cross-disciplinary approaches are needed to answer open questions about the inner workings of GenAI systems.
\end{abstract}

\section{Introduction: Why Generative AI Systems?}

Large Language Models (LLMs) opened doors to the next level of applications of AI systems. For instance, they can serve as assistance in cooking (by generating recipes), writing wishes, and preparing a plan for a presentation, but also allowing communication with machines using natural language. LLMs greatly contributed to a technological jump but also to the way we think about AI. The recent developments in Generative AI (GenAI) like Vision-Language Models \cite{bordes2024introduction}, e.g., in the GPT family (GPT-4V: GPT-4 with Vision \cite{2023GPT4VisionSC}) or in the Google-based models (e.g., Gemini \cite{team2023gemini}), brought multimodal learning in Deep Learning (DL) back and showed great promise in using LLMs as multimodal systems. As a result, there is an increasing interest in multimodal models with LLMs as their core capable of processing various data modalities (not only text but also images, audio, and video) and generating different data types. The rise of multimodal LLMs, or rather multimodal GenAI, requires a new perspective on AI as systems rather than LLMs alone. In such Generative AI Systems (or GenAISys for short), natural language constitutes communication means, instead of pre-defined formal protocols, and modality encoders play the role of I/O interfaces for processing various data sources. Additionally, GenAISys are equipped with databases (or/and knowledge graphs) and external specialized tools, such as a calculator, an app for finding a route from A to B, that communicate with the system through a module for information retrieval and storage. In GenAISys, GenAI models play crucial roles but they are components rather than stand-alone modules. An example of such a GenAI system is a tool-augmented Seq2Seq model proposed by \cite{parisi2022talm} where a fine-tuned T5 LLM is used as a backbone to deal with inputs and outputs to and from some external tools. Another specific use of an LLM as a core of a GenAISys was outlined in \cite{bran2023augmenting} where an LLM communicates with external tools and databases for providing reliable answers for chemistry questions and drawing molecules, among other features.

The increasing complexity of GenAI systems forces us to look at them from a different perspective and ask about their properties as systems instead of single modules. By analyzing \textit{compositionality} \cite{bereska2024mechanistic, cohen2022towards, swan2022road, tripakis2016compositionality} of GenAISys, one can design reliable and verifiable (to some degree, at least) systems. Moreover, studying compositionality and other aspects of GenAISys can reveal their interesting traits or indicate their potential pitfalls, similar to control theory, or more broadly, systems science and engineering \cite{bubnicki2005modern}. In this paper, we want to start the discussion and highlight potential new research directions in the field of Generative AI Systems. Specifically, we are interested in the following questions:
\begin{itemize}
	\item How does the design of GenAISys differ from other AI systems like DL systems?
	\item How to build GenAISys?
	\item How to train GenAISys?
	\item What can we learn from the system-based perspective so that we can build and train \textit{better} GenAISys?
\end{itemize}

We believe that this paper sparks interest among solution architects and practitioners, but also brings theoreticians who could shed new light on Generative AI Systems. There are multiple aspects of designing new GenAI systems that require a more holistic approach. Moreover, there are many open questions about the inner workings of GenAI systems, and answering them requires cross-disciplinary approaches.

%

\section{Compositionality: The main principle behind building systems}

Systems-based analysis seeks to understand complex systems by breaking them into manageable and well-defined subsystems. Similarly to the divide-and-conquer strategy, the premise is that defining and understanding smaller components leads to predictable behavior of the whole, and allows formulation of general designing principles. Moreover, the systems-based perspective opens an opportunity for analyzing the verifiability and reliability of complex systems. These aspects become crucial in contemporary Generative AI systems that go beyond individual modules or even hierarchical structures of neural networks.

Before we move further, we first look into the definition of a \textit{system} \cite{tripakis2016compositionality}. In general, we define an \textit{atomic system} by its \textit{state} and \textit{dynamics} (i.e., an evolution of the state). A \textit{composite system} consists of a set of atomic systems and other composite systems and a set of rules defining how subsystems interact (i.e., \textit{composition}). Then a \textit{system} is either an atomic system or a composite system. The crucial part of composite systems is their \textit{composition}. The basic rule of composition is \textit{compatibility} which determines whether outputs of one subsystem are legal as inputs of another system. By \textit{legal} we mean if they fit syntactically (e.g., the same data type, the same physical units) and semantically (e.g., variables representing the same quantities).

\begin{figure}[htbp!]
	\centering
	\includegraphics[width=0.85\textwidth]{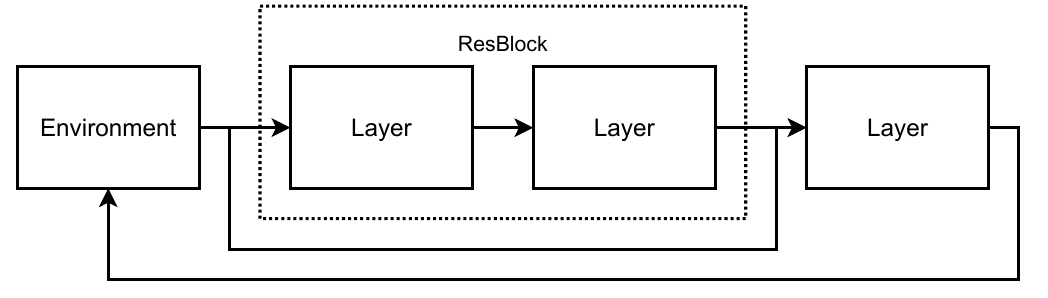}
	\caption{An example of a Deep Learning system: An environment connected with a neural network consisting of three layers.}
	\label{fig:compositionality_example}
\end{figure}

As an example, we can consider neural networks like in Figure \ref{fig:compositionality_example}. Here, we have three layers (two of them organized as a ResBlock) constituting our system (i.e., a neural network) that obtain some inputs from an environment and return outputs. The state of this system is defined by feature maps and dynamics depend on the environment. In this example, we can indicate all our definitions introduced so far. A single layer could be seen as an atomic system with inputs and outputs. The environment alone is a composite system. The dimensionality of inputs and outputs of all layers must be compatible, namely, they must fit with each other. Moreover, the output of the environment must be appropriate for the first layer, and the output of the last layer must be correct for the environment, i.e., they need to correspond to the same quantities (e.g., images).

\begin{figure}[htbp!]
	\centering
	\includegraphics[width=1\textwidth]{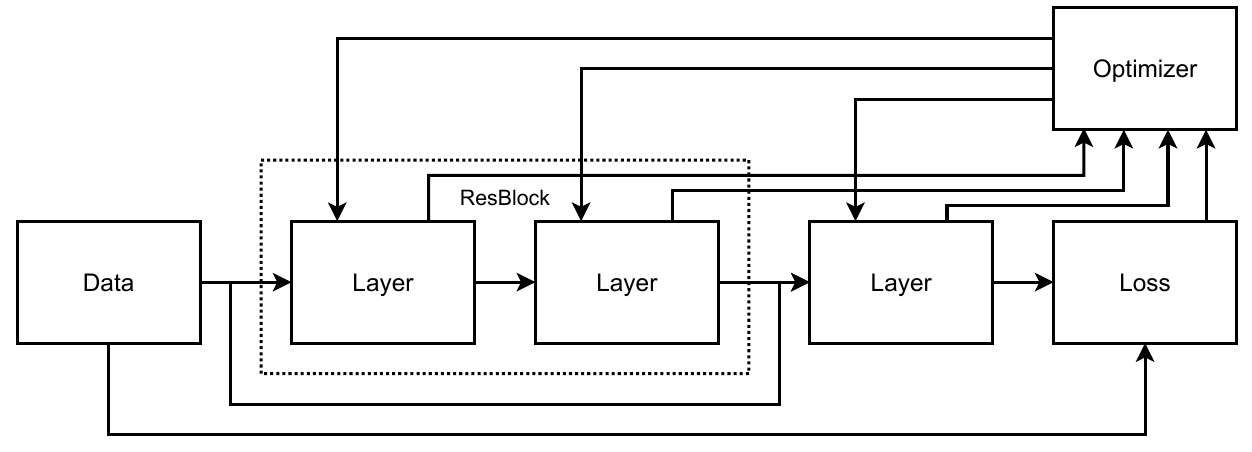}
	\caption{An example of a learning system with a DL model: A neural network (three layers) is connected with a data source, a loss function and an optimizer.}
	\label{fig:compositionality_learning}
\end{figure}

The state of a system can represent different quantities, depending on a considered situation. If we take the neural network in Figure \ref{fig:compositionality_example} but now consider it in a training system, we can get a system like in Figure \ref{fig:compositionality_learning} where we replace the environment with a data distribution, and we add a loss function and an optimizer. In such a system, the state of a neural network corresponds to weights, and dynamics is a training process. Additionally, the arrows connecting an optimizer and the layers represent weight updates. As a result, the interpretation of layer atomic systems is different than the corresponding blocks in Figure \ref{fig:compositionality_example}.

These two examples indicate very important differences but also challenges. First, the diagrams in Figure \ref{fig:compositionality_example} and \ref{fig:compositionality_learning} are not self-explanatory. They require additional descriptions or standards determining the meanings of arrows and blocks. Second, it is important to define the goal of a system. While Figure \ref{fig:compositionality_example} depicts a system at its inference stage, Figure \ref{fig:compositionality_learning} represents a system at its training stage. The goals in both cases are different, hence, the state and the dynamics are defined differently in the two cases. However, all these aspects show that the systems perspective is helpful because we can work with these two examples of composite systems in various manners. For instance, we can design new systems by either replacing some blocks or adding some blocks. Further, we can analyze specific blocks by visualizing the states of subsystems or/and dynamics. In other words, we can check specific properties of the whole system by verifying them for subsystems. 

An important aspect of compositionality is \textit{refinment}. The property of refinement is defined as follows: A system $A'$ refines a system $A$ if: (i) all functionality of $A$ is maintained, and (ii) for all legal values handled by $A$, the outputs of $A'$ and $A$ are the same (e.g., at least semantically or syntactically). If we think about it, this property is not necessarily fulfilled by many DL systems, however, it can be met for the whole system. For instance, two classifiers can output exactly the same classes even though they can have totally different architectures. Another example is replacing activation functions, e.g., sigmoid with relu that results in relu covering outputs of sigmoid and more. The question here is whether this property is necessary for GenAISys. It seems that yes since refinement could play an important role for LLMs and their reliability (e.g., replacing some subsystems of an LLM should not result in unpredictable values, or in building safe LLMs). In \cite{dalrymple2024towards}, the systems-based perspective was used to introduce additional blocks in an AI system: (i) a world model providing a mathematical description of how the system affects the environment it is in, (ii) a safety specification which is a mathematical description of what effects are acceptable, and (iii) a verifier outputting an auditable certificate whether the system satisfies the safety specification relative to the world. Following an example in \cite{dalrymple2024towards}, for a problem of programming language translation, a model is an LLM, a specification can define functional correctness, a world model can provide constraints on the inputs to the program and indicates what an attacker can/cannot do. Such a systems-based perspective helps not only add potentially missing (composite) subsystems but also to develop systems with specific properties, e.g., reliability and security like in the given example.

We highlight the role of compositionality in designing systems, including Generative AI systems. The concept of compositionality as stated here is broad and blurry. Compositionality is studied in various domains, e.g., in DL \cite{bereska2024mechanistic}, causality theory \cite{cohen2022towards}, software engineering \cite{tripakis2016compositionality}. A mathematical field called category theory \cite{fong2018seven} tries to encompass all these perspectives on compositionality so that it can help to find analogies and build upon them. There are multiple great examples of how category theory can be used in various domains, e.g., for understanding causality \cite{cohen2022towards}, or analyzing systems \cite{bakirtzis2021categorical, lloyd2021category}.

\section{Generative AI Systems}

\subsection{LLMs as systems}

Language Models are probabilistic models of a language, e.g., a natural language or a programming language. As such, they take processed language (e.g., n-grams, a bag of words) as input and generate new content. Large Language Models (LLMs) are language models parameterized by neural networks, e.g., Recurrent Neural Networks (RNNs), transformers, or state space models. However, in fact, LLMs are not only language models, they consist of multiple modules. Each LLM requires a \textit{tokenizer} to turn text into numbers (e.g., integers), and an \textit{embedding} that changes tokenized text to real-valued vectors. Sometimes, both modules are treated as one (e.g., vectorizers in \texttt{scikit-learn}). A popular choice for a tokenizer these days is \textit{byte pair encoding} which greedily merges commonly occurring sub-strings based on their frequency \cite{gage1994new}. The embedding module serves only a single purpose, namely, to map a token represented as a one-hot vector to a real-valued vector of size $D$. Then, after processing embeddings using a neural network, the output must be de-tokenized to a string again.

\begin{figure}[htbp!]
    \centering
    \begin{tabular}{cc}
         \includegraphics[width=0.45\textwidth]{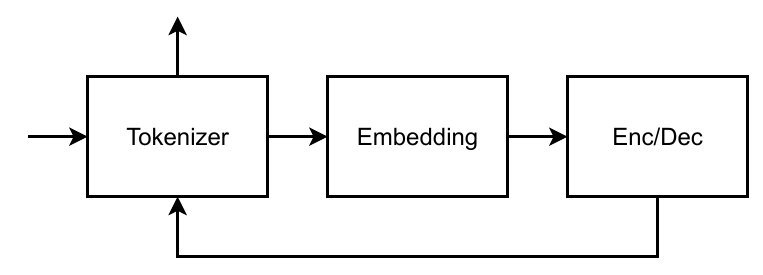} & \includegraphics[width=0.45\textwidth]{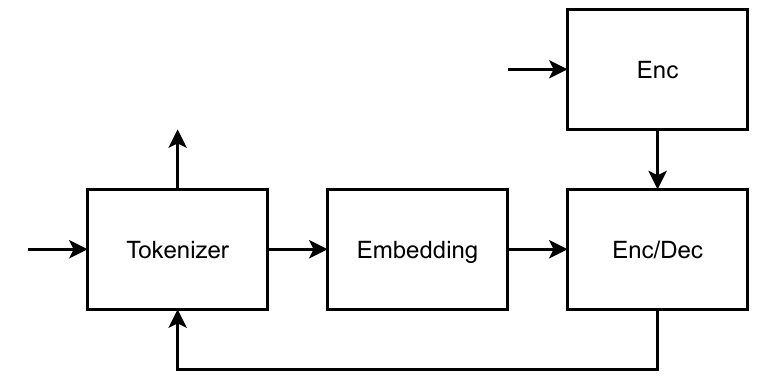}\\
         \textbf{A} & \textbf{B}
    \end{tabular}
    \caption{Diagrams for LLMs: \textbf{A.} An unconditional LLM. \textbf{B.} A conditional LLM.}
    \label{fig:llms_diagrams}
\end{figure}

In general, we can distinguish three types of LLMs:
\begin{enumerate}
    \item \textbf{Encoders} take a piece of text (string) and return an encoding, i.e., a numerical representation of the input. Encoders can have access to the whole input at any point of processing and they do not require any specific constraints. They provide outputs in a single forward run both during training and at the inference time.
    \item \textbf{Decoders} are used for \textit{generating} new texts (strings). They can be seen as autoregressive models and, as such, neural networks parameterizing them must be \textit{causal}. For decoders, the sampling procedure is an iterative process, which is typically slow.
    \item \textbf{Encoder-Decoders} and \textbf{Encoder-Encoders} are LLMs that are conditioned on additional information. Therefore, an additional encoder is used to process conditioning, and then an encoder or a decoder provides an encoding of input text or generates new text, respectively.
\end{enumerate}

In Figure \ref{fig:llms_diagrams}.A, you can see a schematic representation of an encoder or a decoder, and in Figure \ref{fig:llms_diagrams}.B there is an encoder-encoder(decoder) presented. We highlight various blocks on purpose to highlight subsystems and their composition. Like in previous examples (see Figure \ref{fig:compositionality_example} \& \ref{fig:compositionality_learning}), compatibility of inputs and outputs is important, and the overall composition determines the gradient/information flow during training/inference.

\subsection{Going beyond LLMs: Building Generative AI Systems (GenAISys)}

Perceiving LLMs as systems is helpful to understand their building blocks and to design them for specific applications. This is also constructive to realize that LLMs could be extended and even generalized. For instance, Vision-Language Models (VLMs) process both visual inputs and textual inputs and return task-specific outputs. The idea of multimodal processing and learning is one of the most important landmarks towards fully-capable AI systems, and Generative AI models seem to play the key role in this. Analogously to LLMs, encoders process various data modalities for further use by a generative model (e.g., an LLM or a compute vision model). We can extend such systems with external tools like calculators or algorithms for planning to further expand their functionality. As a result, we design Generative AI Systems with a generative AI model in the center, data encoders and external tools (incl. databases). Overall, a GenAISys consists of the following parts:
\begin{itemize}
    \item \textbf{Data encoders (DEs)}: All input (raw) data are processed using models that encode them into tensors. Typically, these encoders are pre-trained and kept non-trained (\textit{frozen}). We can use any foundation model (FM) as an encoder, e.g., BERT for text encoding, ConvNeXT (without the predictive head) for image encoding, etc. Encoders could be composed of other modules (encoders) as well. For instance, speech could be transformed into text (e.g., using Whisper) and then text could be represented as a tensor (e.g., using BERT). Data encoders play a crucial role in formulating GenAISys and their introduction was a catalyzer for the development of GenAI systems as we know it now.
    \item \textbf{GenAI Model (GeM)}: The \textit{central unit} of a GenAISys is a GenAI model (e.g., an LLM) that processes (encoded) input data, communicates with external \textit{memory} (a database) and external tools, and generates an output. It is the core part of the whole architecture. It can have its own short-term memory (a cache), e.g., in the form of available input tokens, and built-in instructions for communicating with other components. Moreover, it can be equipped with other (sub)models carrying out other tasks, e.g., Named Entity Recognition (NER) for detecting specific instructions in inputs. In the simplest form, it could be trained in such a way that specific instructions in natural language trigger specific actions like running a calculator. Overall, a GeM can be a complicated system by itself, comprising multiple modules including separate models.
    \item \textbf{Retrieval/Storage module (R/S)}: This module serves an extremely important function to assist GeM with retrieving facts (\textit{long-term memory}) but also utilizing inputs processed by DEs. Additionally, this module can have its own models (e.g., BERT to embed a part of the input text of an internal instruction) and sets of instructions (e.g., routing algorithms).
\end{itemize}

\begin{figure}[!htbp]
    \centering
    \includegraphics[width=0.65\textwidth]{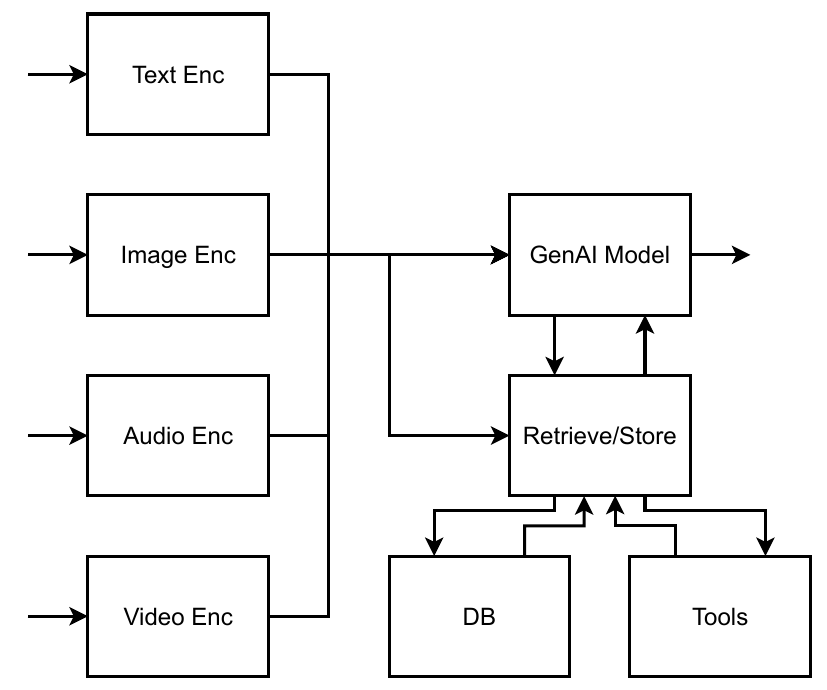}
    \caption{A general architecture of a Generative AI system with encoders for various modalities, a retrieval/storage module for accessing external tools and databases, and a central generative AI model producing new content (output). The snowflake icon represents that a module is "frozen" (i.e., already trained).}
    \label{fig:genaisys_complex}
\end{figure}

A general scheme of GenAISys is presented in Figure \ref{fig:genaisys_complex}. Please keep in mind that this is a simplified architecture that highlights only the main components. Each block could consist of multiple subsystems like models, and procedures (algorithms). Note that the diagram looks like the general computer architecture. After all, such architectures are quite \textit{natural}. However, unlike computer architectures, GenAISys consists of trainable components and, eventually, is extremely flexible in the sense of its functionality. We want to highlight that GenAISys could be further composed with other systems. For instance, with a safe system proposed in \cite{dalrymple2024towards}.

\subsection{Training GenAISys as systems}

Training of a GenAISys is non-trivial since all components are trainable and the whole system consists of neural networks with millions if not billions of weights. 
As a result, training such systems end-to-end is infeasible for currently available hardware. Hopefully, in the future, with new hardware development, and new training schemes, GenAISys will be trained in a better way and, eventually, will improve through the utilization of multiple data modalities at the same time.

\begin{figure}[!htbp]
    \centering
    \includegraphics[width=1\textwidth]{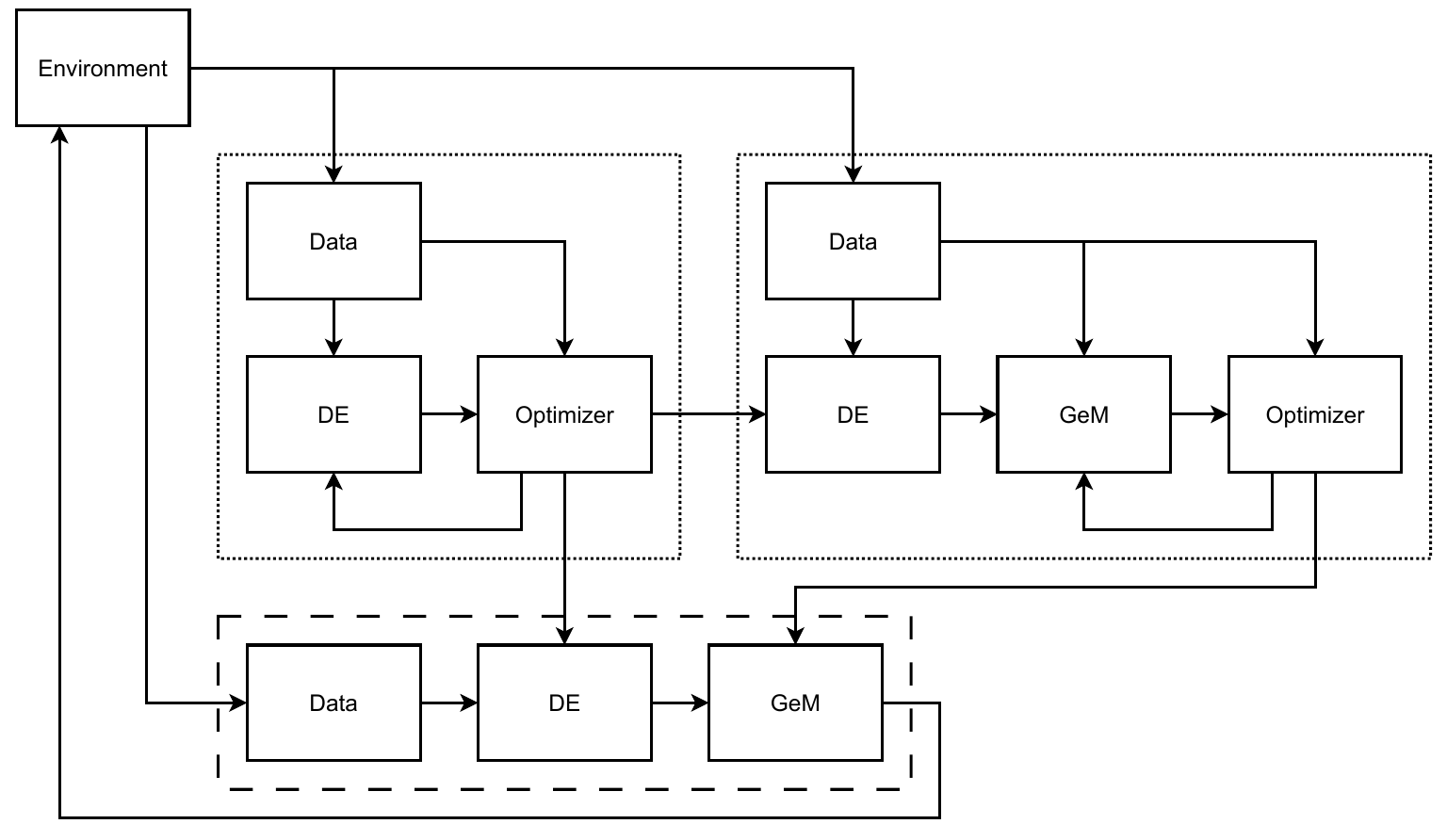}
    \caption{An example of a learning system with GenAISys: Data is taken from the environment and it is used to first pre-train a DE (the first dotted subsystem) and subsequently a GeM is fine-tuned (the second dotted subsystem). Eventually, both trained subsystems are used for inference (the dashed subsystem).}
    \label{fig:genaisys_learning}
\end{figure}

Nowadays, a widely applied solution is to take advantage of foundation models \cite{bommasani2021opportunities} that are pre-trained separately to formulate DEs and an R/S subsystem. Then, by \textit{freezing} these components, a GeM is trained. As a result, we distinguish two steps:
\begin{enumerate}
    \item \textbf{Pre-training}: In the initial stage, data encoders and/or other subsystems (e.g., an R/S subsystem) are trained involving large volumes of data. The goal of this stage is to train general patterns in data, e.g., grammar and co-occurrences of words, a specific programming language, and a representation of images.
    \item \textbf{Fine-tuning}: Pre-trained models are later kept frozen and are used for processing data for a GeM. Alternatively, they can be further specialized on another dataset for a downstream task. For instance, an LLM can be trained on specific data, e.g., legal data to generate legal documents or a new programming language. However, the LLM can be also fine-tuned to carry out other tasks like text summarization, Q\&A, text classification, sentiment analysis, etc. Eventually, once all subsystems are trained, the GeM is trained with frozen DEs and an R/S module.
\end{enumerate}
An example of a learning system with GenAISys is depicted in Figure \ref{fig:genaisys_learning}. Note that the states of the dotted subsystems and the dashed subsystems are different.

These two steps are quite general and fully depend on a given task at hand. For instance, the first Generative Pretrained Transformers (GPTs) \cite{radford2018improving} were pre-trained using the negative log-likelihood or they were initialized by training with the masked loss like in Bidirectional Encoder Representations from Transformers (BERT) \cite{devlin2018bert}. Eventually, GPTs were fine-tuned with the negative log-likelihood loss. It is also possible to combine various losses to better reflect the tasks at hand. This idea was utilized in pre-training LLMs for various problems at once \cite{dong2019unified} or for pre-training an LLM for molecules \cite{izdebski2023novo}.

The problem with fine-tuning DEs and/or GeMs lies in their size. Ideally, fine-tuning should be quick and cheap but it is hard to achieve if we deal with models with billions of weights. A possible solution is to use one of the techniques known as Parameter-Efficient Fine-Tuning methods (PEFT) \cite{peft} like Low-Rank Adaptation (LoRA) \cite{hu2021lora}. The general idea lies in adding low-rank matrices to frozen pre-trained weights and fine-tuning only these low-rank matrices. As a result, after applying LoRA, the additional overhead is at a level of $1-5\%$ of the original number of weights.

\subsection{Examples of GenAISys}

Following the general scheme for GenAISys in Figure \ref{fig:genaisys_complex}, we can indicate how currently used Generative AI approaches fit this scheme. We focus on Large Vision Models and a specific LLM-based solution for reliable text generation. There are other examples and we present an arbitrary subset of those, however, we believe they properly present the ideas presented in this paper.

\subsubsection{RAGs} The main drawback of decoder-based LLMs is hallucinating when a prompt takes the model away from its "comfort zone" (i.e., from regions where training data lie in the representation space). This could be fixed to some degree with proper fine-tuning, however, LLMs tend to make up or skip some facts. Since their responses are typically very colorful with distinctive and unusual wording, human beings can miss some deficiencies and false statements. In many applications, there is no place for fake news or, in general, outcomes cannot be untrustworthy. For instance, in health-related situations like medicine, drug discovery, or manufacturing (e.g., diagnostics), there is no space for made-up facts. Therefore, even though generative LLMs are so popular, they do not pose a possible solution due to their high risk of hallucinating. 

\begin{figure}[!htbp]
    \centering
    \includegraphics[width=0.5\textwidth]{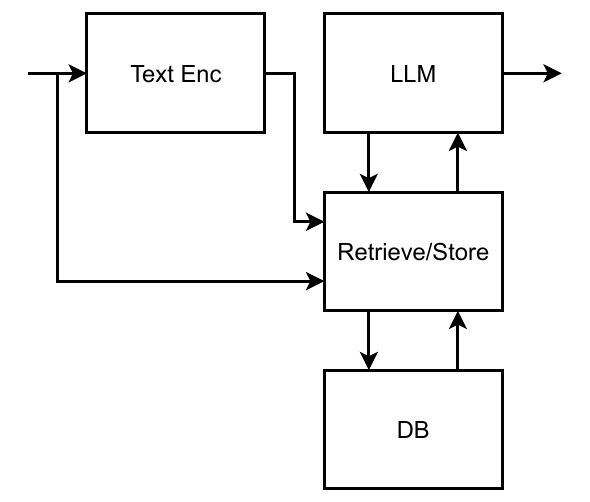}
    \caption{A schematic representation of an Retrieval-Augmented-Generation (RAG) architecture.}
    \label{fig:genaisys_rag}
\end{figure}

A huge breakthrough, especially in real-life applications, came with Retrieval Augmented Generations (RAGs) \cite{lewis2020retrieval}. The idea is based on utilizing two LLMs (an encoder-LLM and a decoder-LLM) and a database of texts (facts). The encoder-LLM is used in two ways: (i) for embedding all texts in the database, and (ii) for embedding an incoming query. For a new query, the closest documents are picked based on the distance between the embedding of the query and the embeddings of the documents in the database. Eventually, the closest documents, together with the query, are passed to the decoder-LLM to generate an outcome. Since the outcome is based on the decoder-LLM and real documents, there is a much lower chance of hallucinations. Moreover, with a bit of tweaking around, the RAG could rely heavily on facts provided during the retrieval stage. 

The diagram for a RAG is presented in Figure \ref{fig:genaisys_rag} and it corresponds very closely to the general scheme of GenAISys in Figure \ref{fig:genaisys_complex} where the decoder-LLM is the GeM, and the encoder-LLM is the DE and it is also used as a part of the R/S module. 

\subsubsection{Speech2Txt} 
\begin{figure}[!htbp]
    \centering
    \includegraphics[width=0.5\textwidth]{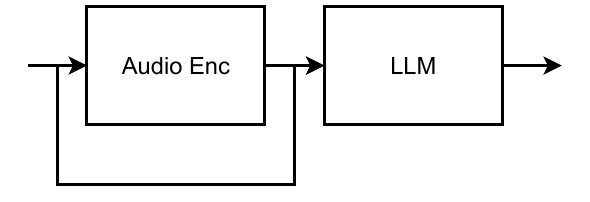}
    \caption{A schematic representation of a Speech2Txt architecture.}
    \label{fig:genaisys_whisper}
\end{figure}

A great example of a GenAISys for transforming speech to text is Whisper \cite{radford2023robust}, an encoder-decoder transformer with a specific form of the encoder that first represents raw speech (audio) using a log-magnitude Mel spectrogram before being fed to an encoder-transformer for processing audio signal and then to a decoder-transformer for generating text. The model is an automatic speech recognition system with 39M weights (a tiny version) to even 1.55B weights (a large version). It was trained on 680,000 hours of multilingual and multi-task supervised data collected from the web. This model achieved SOTA performance at the time of its release and still it remains one of the top Speech2Txt models available. The tiny version could even deployed on edge devices. Whisper is a great example of how GenAISys can be formulated and how important it is to compose various components together for more advanced tasks like automatic speech recognition. The Speech2Txt diagram is presented in Figure \ref{fig:genaisys_whisper}.

\subsubsection{Large Vision Models (LVMs)} 
\begin{figure}[!htbp]
    \centering
    \begin{tabular}{cc}
        \includegraphics[width=0.45\textwidth]{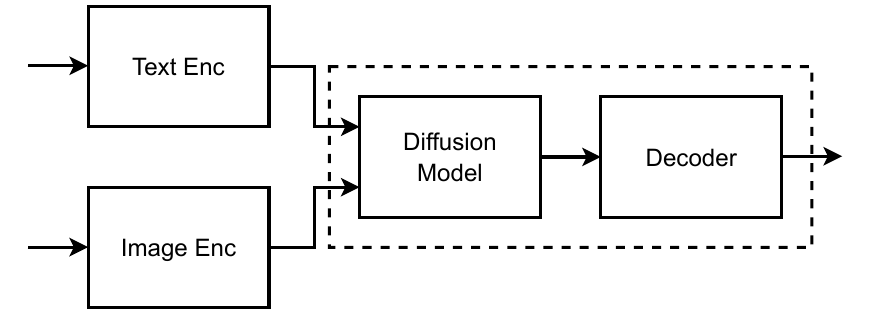} & \includegraphics[width=0.45\textwidth]{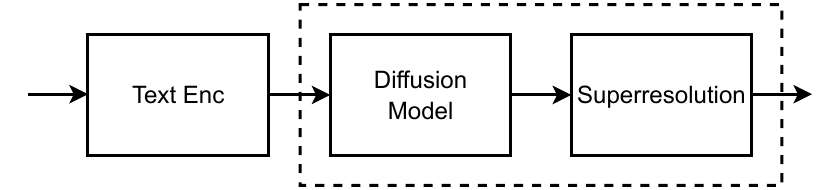} \\
        \textbf{A} & \textbf{B} 
    \end{tabular}
    \caption{Examples of LVMs: \textbf{A.} Stable Diffusion (i.e., latent diffusion). \textbf{B.} ImaGen.}
    \label{fig:genaisys_lvms}
\end{figure}

Beside LLMs, Large Vision Models (LVMs) are perfect examples of GenAISys. There are many models that fall under the umbrella of Img2Img or Img2Txt, but the most popular LVMs these days are Txt2Img. Since the original paper on latent diffusion models \cite{rombach2022high}, the resulting models like Stable Diffusion 2 and very recent Stable Diffusion 3, are widely used for generating images for a given prompt. Latent diffusion models (Stable Diffusion) or Dalle 2 \cite{ramesh2022hierarchical} with a diffusion-based prior fit perfectly a scheme in Figure \ref{fig:genaisys_lvms}.A. Comparing these LVMs to a general GenAISys, a text encoder and an image encoder (for either training or reconstruction) are DEs while a combination of a diffusion model and a decoder is a GeM. These models do not use any database or external tools, however, it is possible to use those to modify images.

Another example of an LVM in the form of Txt2Img is ImaGen \cite{saharia2021image} which uses a T5-based text encoder and a diffusion model together with superresolution blocks. The corresponding architecture is presented in Figure \ref{fig:genaisys_lvms}.B (the superresolution module consists of multiple steps, going from 64x64 images to 1024x1024 images). Again, this is a complex GenAISys even though it is composed of three blocks but it has about 13B weights (11B weights for T5, about 2B weights for a UNet used in the diffusion model and the superresolution module) which is a large model in terms of the number of weights.

\section{Future \& Challenges}

We can look at GenAISys from different perspectives. First, we can think very pragmatically about their functionality and how they can be implemented. For instance, we can use LLMs as a backbone for Operating Systems \cite{packer2023memgpt}. The idea is the following: Various events are parsed to a virtual context that is processed by an LLM and its output is parsed to specific functions. The idea is very appealing since such an OS is trainable, and it communicates with a user in natural language. Moving towards general GenAI-based (operating) systems seems like the future, and the next step of cloud-based systems. Indeed, GenAISys can be deployed locally, but also in a cloud server; or as a hybrid (e.g., a GeM, a cache storage, and DEs are local but external tools and storage are in a cloud). The last option can be especially appealing for manufacturing since all real-life operations must be executed in real-time while data storage and other operations are carried out by external services (or agents).

Nowadays, Generative AI meets Responsive AI results in Agentic AI, i.e., the development of GenAISys-based agents operating autonomously. This idea is pushed by many Big Tech players. For instance, Microsoft proposed a framework for conversational LLM-based agents called AutoGen \cite{wu2023autogen}. OpenAI sees their chatbots (incl. ChatGPT) as agents and by equipping them with various tools and features, they could serve as \textit{co-pilots} (i.e., assisting a human operator by proposing partial solutions) or \textit{auto-pilots} (i.e., assisting human operators by proposing complete solutions). The analogy here corresponds to controlling a plane and a co-pilot helps to stabilize a flight while an auto-pilot takes care of flying. In both cases though a human pilot can take over at any point.

Second, we can focus on the design principles of GenAISys. Instead of building applications using experience, we should develop design patterns that would benefit the work of solution architects. GenAISys are very specific since they consist of billions of weights and their training is challenging. As a result, diagrams like in Figure \ref{fig:genaisys_learning} are needed. Additionally, we may require new types of modeling languages that would meet the requirements of GenAISys.

Third, we have deep learning libraries that can be used for building GenAISys. However, we need new specialized frameworks (including GUI) that can speed up working with and building GenAISys. Since composing GenAISys could be accomplished by combining specific blocks together, there is no problem in creating low-code programming tools. Eventually, we could expect visual programming environments similar to Simulink but for GenAISys.

Fourth, a natural question is whether GenAISys could be better formalized (e.g., using category theory \cite{fong2018seven, swan2022road}) and whether such a formal perspective could help formulate new systems. It is important to have formal tools to analyze systems for such aspects as reliability, safety, and robustness. 

Overall, the future seems to belong to (deep) generative modeling and Generative AI Systems that are inevitable next steps in the evolution of AI. They will assist in many jobs, ranging from office jobs, healthcare, and education to the industry like manufacturing. There are many other aspects like embodied AI \cite{roy2021machine} or cyber-physical systems (with humans) \cite{tripakis2016compositionality}, but still GenAISys will be necessary to formulate an artificial "brain" and/or GenAI-based agents and apps. 

\subsubsection*{Acknowledgments}
The author would like to express his high appreciation to Prof. Zdzis\l aw Bubnicki (R.I.P) and Prof. Jerzy \'{S}wi\k{a}tek who made a great impact on the author and his systems-based perspective on AI. Moreover, the author is thankful to Dr. Efstratios Gavves for many discussions about various aspects of AI, including AI systems.

\bibliography{tmlr}
\bibliographystyle{tmlr}


\end{document}

%% file: math_commands.tex
\usepackage{amsmath,amsfonts,bm}

\def\eqref#1{equation~\ref{#1}}

\def\1{\bm{1}}

\DeclareMathAlphabet{\mathsfit}{\encodingdefault}{\sfdefault}{m}{sl}
\SetMathAlphabet{\mathsfit}{bold}{\encodingdefault}{\sfdefault}{bx}{n}

